\documentclass[10pt,twocolumn,letterpaper]{article}

\pdfoutput=1

\usepackage{cvpr}
\usepackage{times}
\usepackage{epsfig}
\usepackage{graphicx}
\usepackage{amsmath}
\usepackage{amssymb}
\usepackage{adjustbox}

\usepackage[pagebackref=true,breaklinks=true,letterpaper=true,colorlinks,bookmarks=false]{hyperref}

\usepackage{gensymb}
\usepackage{algorithm2e}
\usepackage{ upgreek }
\usepackage{multirow}
\usepackage{makecell}
\usepackage{diagbox}
\usepackage{pifont}
\usepackage{subcaption}
\usepackage{authblk}

\newcolumntype{P}[1]{>{\centering\arraybackslash}p{#1}}
\DeclareMathOperator*{\argmax}{argmax} 
\DeclareMathOperator*{\argmin}{argmin} 

\cvprfinalcopy 

\def\cvprPaperID{1232} 

\ifcvprfinal\fi
\begin{document}

\title{A Large Dataset for Improving Patch Matching}

\author[1]{Rahul Mitra}
\author[1]{Nehal Doiphode}
\author[1]{Utkarsh Goutam}
\author[2]{Sanath Narayan}
\author[2]{Shuaib Ahmed}
\author[1]{Sharat Chandran}
\author[1]{Arjun Jain}
\affil[1]{Indian Institute of Technology Bombay, Mumbai, India}
\affil[2]{Mercedes Benz Research and Development India, Bengaluru, India}
\affil[ ]{\tt\small \{rmitter,sharat,ajain\}@cse.iitb.ac.in, nehald@ee.iitb.ac.in, utkarshg@iitb.ac.in, \{sanath.narayan,shuaib.ahmed\}@daimler.com}

\renewcommand\Authands{ and }


\maketitle

\begin{abstract}

We propose a new dataset for learning local image descriptors which can be used for significantly improved patch matching. Our proposed dataset consists of an order of magnitude more number of scenes, images, and positive and negative correspondences compared to the currently available Multi-View Stereo (MVS) dataset from Brown \etal~\cite{mvs-new}. The new dataset also has better coverage of the overall viewpoint, scale, and lighting changes in comparison to the MVS dataset. Our dataset also provides supplementary information like RGB patches with scale and rotations values, and intrinsic and extrinsic camera parameters which as shown later can be used to customize training data as per application. 


We train an existing state-of-the-art model on our dataset and evaluate on publicly available benchmarks such as HPatches dataset~\cite{hpatches} and Strecha et al.~\cite{strecha} to quantify the image descriptor performance. Experimental evaluations show that the descriptors trained using our proposed dataset outperform the current state-of-the-art descriptors trained on MVS by $8\%$, $4\%$ and $10\%$ on matching, verification and retrieval tasks respectively on the HPatches dataset. Similarly on the Strecha dataset, we see an improvement of $3$-$5\%$ for the matching task in non-planar scenes.

\end{abstract}

\section{Introduction}{\label{sec-intro}}

Finding correspondences between images using descriptors is important in many computer vision tasks such as 3D reconstruction, structure from motion (SFM) ~\cite{photo-tourism}, wide-baseline matching~\cite{strecha},
stitching image panoramas~\cite{panorama}, and tracking~\cite{he2009surf, surf}. However, due to changes in viewpoints, scale variations, occlusion, variations in illumination, and shading in the real world scenarios, finding correspondences in-the-wild is challenging and it is an active area of research.

Traditionally, handcrafted descriptors such as SIFT ~\cite{sift}, SURF~\cite{surf}, LIOP~\cite{liop} were used. These type of descriptors encode pixel, super-pixel or sub-pixel level statistics. However, handcrafted features do not have ability to capture higher structural level information. On the other hand, learning based descriptors using Convolutional Neural Networks (CNNs) have the potential to capture higher level structural information and also to generalize well. Hence, CNN based descriptors are gaining more importance in recent years~\cite{deepdesc, deepcompare, matchnet, tfeat, l2-net, hardnet,spread-out}.

Many research works using CNN based descriptors, focus on the architecture~\cite{deepcompare}, defining better loss function~\cite{tfeat,spread-out}, and improving training strategies \cite{l2-net,hardnet} to enhance the quality and achieve state-of-the-art results. As noted in~\cite{hpatches}, it is unclear that these descriptors can be used for applications where data is not representative of the dataset they are trained with. This is because few datasets are small~\cite{generated-dataset, oxford}, few lack diversity~\cite{mvs-new, DTU}, and in few datasets scenes are obtained through controlled laboratory experiments using small toys~\cite{DTU}. As a result, despite a wide variety of datasets being available in the literature~\cite{mvs-new, DTU, oxford, generated-dataset,CDVS}, they cannot be employed to design descriptors for applications in-the-wild.

Recently, Hpatches dataset~\cite{hpatches} has been proposed as a benchmark for evaluation of local features. This dataset is large and diverse with clear protocols for evaluation metrics and reproducibility . Hpatches dataset has overcome the shortcomings of older smaller datasets such as Oxford-Affine~\cite{oxford} that were used as evaluation benchmarks. Although Hpatches dataset is an excellent benchmark for evaluation, this dataset is seldom used for training as the images in its scenes are related only by 2D homography and such assumptions cannot be made for real-world applications.

Frequently used dataset for training and learning local descriptors is the Multi-View Stereo (MVS) dataset from Brown \etal~\cite{mvs-new}. The MVS dataset comprises of matching and non-matching pairs for training obtained from scenes of real world objects captured at different viewpoints. However, MVS dataset consists of only 3 scenes and cannot be considered as diverse enough. Data augmentation is one of the traditional method employed to increase the size of dataset. Mishchuk \etal~\cite{hardnet} highlighted the importance of data augmentation and achieved state-of-the-art results. Regardless, data augmentation cannot substitute the advantages of training with a larger and diverse dataset. These drawbacks of the current datasets limit the potential of powerful CNN based approaches and highlight the necessity for an improved, next generation dataset as concluded in~\cite{local-feature-eval}.



In this paper, we introduce a novel dataset for training CNN based descriptors that overcomes many drawbacks of current datasets such as MVS. It has sufficiently large number of scenes, is diverse, and has better coverage of the overall viewpoint, scale, and illumination. Moreover, this dataset contains RGB patches including information such as location, scale, and rotation to reverse map them onto the scene. Additionally, this dataset also has intrinsic and extrinsic camera parameters for all the images in a scene which enables one to incorporate the functionality of setting scale and viewpoint variations for matching correspondences. With all the ingredients, this dataset is conducive and ideal for learning descriptors which can also be customized to various diverse tasks of learning including narrow base line matching and wide baseline matching.

A sampling technique for generating matching correspondences is also introduced. This type of sampling ensures that the training dataset has sufficient variations in viewpoint and scale while generating patch-pairs and avoids the generation of redundant patch-pairs having similar contextual information.

We use the current state-of-the-art Hardnet model~\cite{hardnet} and train using the proposed dataset, while maintaining the training strategy identical to~\cite{hardnet}. We show that, using our dataset for training improves the performance over Hardnet for various tasks on different benchmark datasets.


\section{Related Work}{\label{section-related}}

\begin{figure}[h!]
\includegraphics[width=\linewidth]{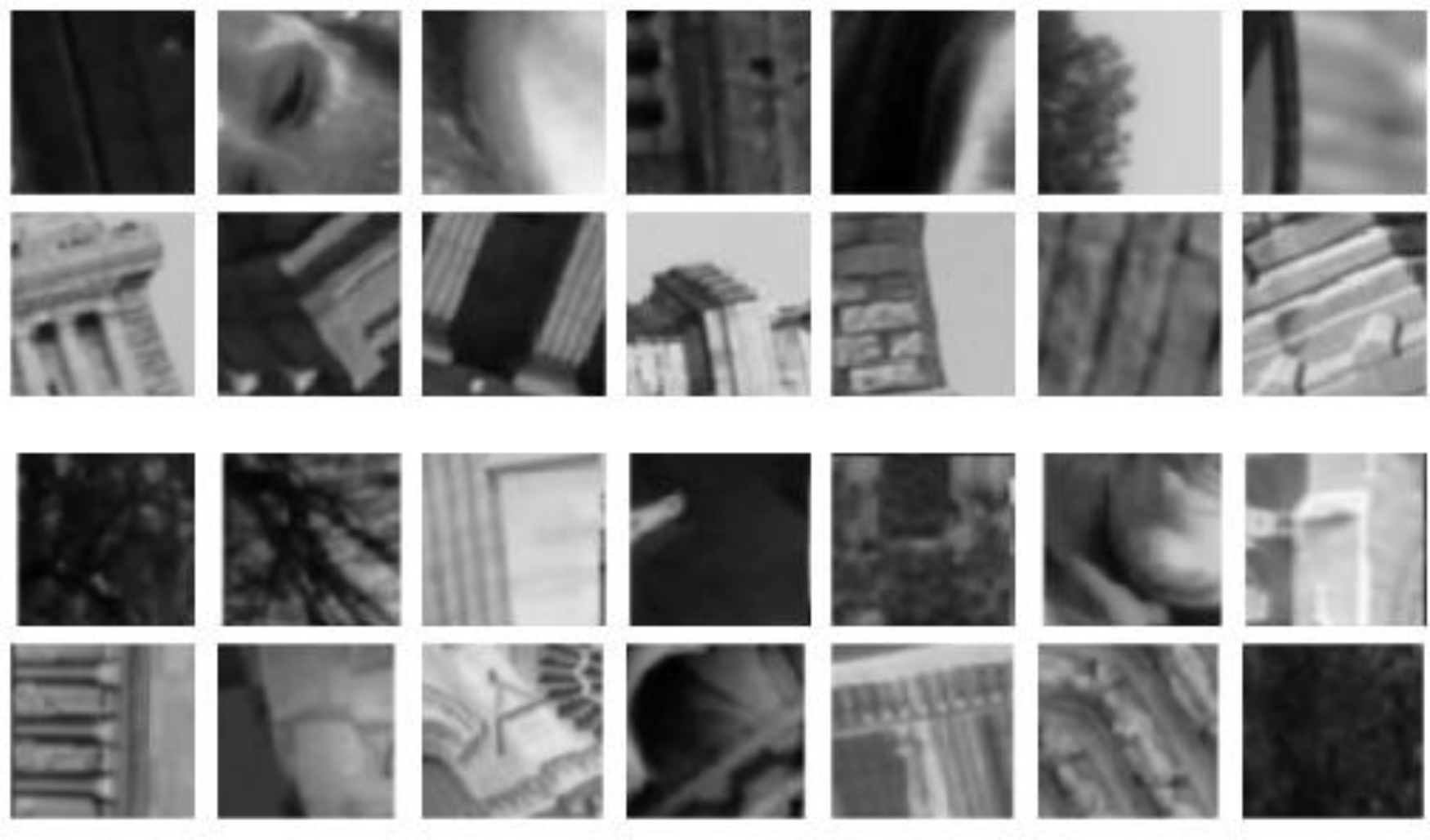}
\caption{The top two rows show incorrect matching pairs from the \emph{liberty} scene. Patches in a column form a pair. The bottom two rows shows the same for \emph{notredame}}
\label{mvs-noise}
\end{figure}

The success of CNNs in various computer vision tasks can be partly attributed to availability of large datasets for training. An ideal dataset for learning a particular task should capture the all the real world scenario involved with the task. An example being the ImageNet~\cite{image-net} dataset for image classification. In the context of learning patch descriptors the dataset provided by Brown et al.~\cite{mvs-new} is the most widely used for training. The dataset contains 3 scenes \emph{viz}., \emph{liberty}, \emph{notredame} and \emph{yosemite}. Each scene consists of a large collection of images. Dense 3D point cloud and visibility maps are estimated from the set of images. The 3D points are projected in different reference images accounting visibility to extract patches. Each scene contains more than 400,000 patches. Patches belonging to same 3D point form matching pairs. However, the dataset suffers from two major drawbacks. Firstly, it lacks data diversity as it contains only 3 scenes. Secondly, inconsistencies in the predicted visibility maps produce noisy matching pairs. In Fig.~\ref{mvs-noise}, few noisy matching pairs from \emph{liberty} and \emph{notredame} scenes are shown. These limitations severely restrict the performance of the descriptors trained with the dataset as shown in Sec.~\ref{sec-results}.

The \emph{DTU} dataset~\cite{DTU} contains images and 3D point clouds of small objects obtained using a robotic arm in a controlled laboratory environment. Images are taken from different view points with varying illumination. Although the size of the dataset is big in number of images, it does not capture intricacies of images in the wild. 

\begin{figure}[h!]
\includegraphics[width=\linewidth]{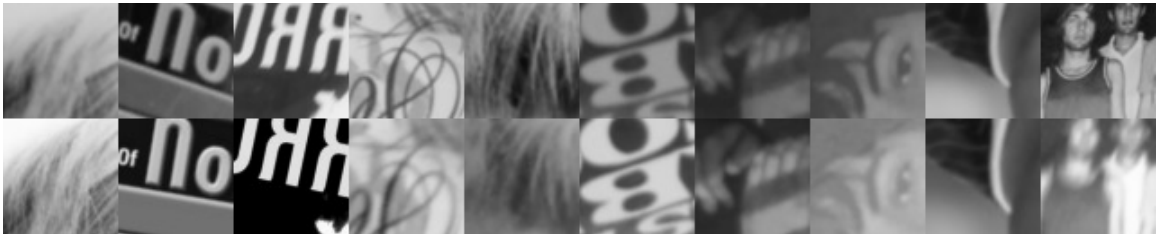}
\caption{Shows sample matching pairs from the CDVS dataset~\cite{CDVS}. It is evident that the pairs do not encompass the necessary challenges encountered in the wild.}
\label{CDVS-pairs}
\end{figure}

The CDVS dataset~\cite{CDVS} is another large patch based dataset offering more number of scenes than the MVS dataset. However, as shown in Fig.~\ref{CDVS-pairs} the matching pairs in the dataset does not have severe deformations. A quantitative analysis depicting the weakness of this dataset is presented in~\cite{tfeat-thesis}. 

The Oxford-Affine dataset~\cite{oxford} is a small dataset containing 8 scenes with sequence of 6 images per scene. The images in a sequence are related by homographies. Although the dataset is suitable for benchmarking evaluations, it is too small for training CNN models. Similar to Oxford-Affine, another dataset exists where matching pairs are created synthetically~\cite{generated-dataset}. In this dataset, every scene contains a reference image and a collection of images which are transformations of the reference image. The dataset has good variations in scene content and deformations. However, the deformations are only limited to homographies. Table~\ref{dataset-comp} gives a comparison of the various publicly available datasets.

\begin{table}[h!]
\centering
 \begin{tabular}{||c | c |c |c |P{3em}||} 
 \hline

\diagbox[width=8em]{Dataset}{Features}
 & Diverse&Real&Large&Non-Planar\\ [0.2ex] 
 \hline\hline
MVS~\cite{mvs-new}         &            & \checkmark & \checkmark & \checkmark\\
 \hline
DTU~\cite{DTU}        &            & \checkmark & \checkmark & \checkmark\\
 \hline
CDVS~\cite{CDVS}        & \checkmark &             & \checkmark & \\
\hline
Oxford-Affine~\cite{oxford} & \checkmark &  \checkmark &           & \\
\hline
Synthetic~\cite{generated-dataset}   & \checkmark &  \checkmark &           & \\
\hline
{\bf Proposed}      & \checkmark & \checkmark & \checkmark & \checkmark\\
\hline

\end{tabular}
\vspace{2mm}
\caption{\label{dataset-comp} Shows a comparison of different features of various publicly available datasets. It is evident that while our proposed dataset incorporates all the features, others exhibit a subset of it.}
\end{table}



Mishchuk \etal~\cite{hardnet} used the MVS dataset for training their network and noted that the state-of-the-art results can be achieved by using better CNN architectures and training procedures. However, Schoenberger \etal~\cite{local-feature-eval}, through extensive experiments, highlighted the importance and the necessity of a better training dataset for learning patch descriptors.  


Based on all these considerations, the contributions of the paper are:\\
(a). A large and novel PS dataset for learning patch descriptors, created from real-world photo-collections, having a good coverage of viewpoint, scale and illumination is proposed. \\
(b). A sampling technique to generate high quality matching correspondences without resulting in redundant patch matches is proposed. \\
(c). By training the current state-of-the-art model on the proposed dataset and outperforming the model, we show that alongside having better models and training procedures, the quality of the training dataset is also important in realizing the potential of the CNN.


\section{Proposed PS Dataset}{\label{section-dataset}}

The dataset proposed in this paper is called PhotoSynth (PS) dataset as images were collected by crawling through Microsoft PhotoSynth. This section focuses on various aspects of the dataset. The description about the scenes and images collected to form the dataset is detailed in Sec.~\ref{section-PS-qual} followed by the methodology adopted to create data for learning local descriptors out of the vast collection of images and the format of dataset in Sec.~\ref{sec-create-data} and Sec.~\ref{sec-sampling}

\begin{figure}[t]
\begin{subfigure}[b]{\linewidth}
\includegraphics[width=\textwidth]{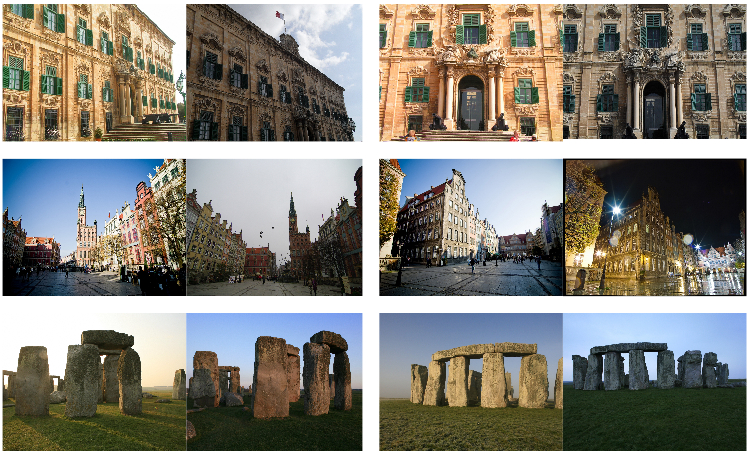}
\caption{Image pairs showing illumination variation.}
\end{subfigure} \\

\begin{subfigure}[b]{\linewidth}
\includegraphics[width=\textwidth]{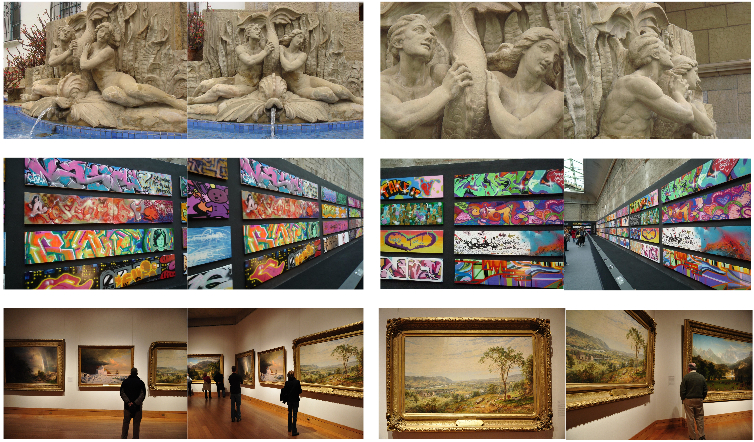}
\caption{Image pairs showing viewpoint variation.}
\end{subfigure} \\

\begin{subfigure}[b]{\linewidth}
\includegraphics[width=\textwidth]{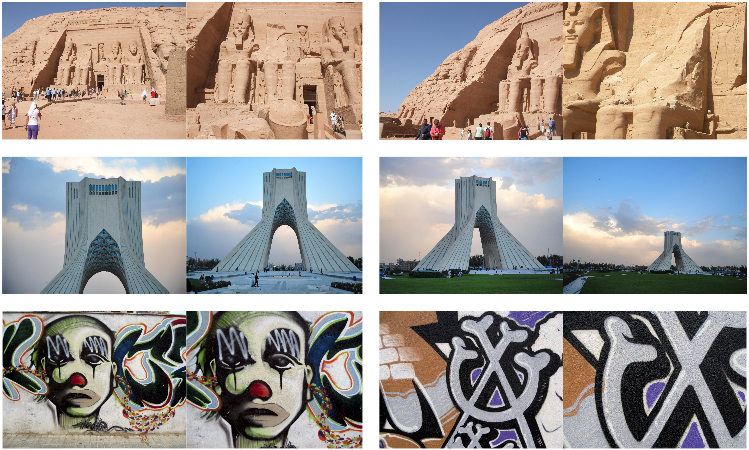}
\caption{Image pairs showing scale variation.}
\end{subfigure}
\caption{\label{ps-sample} Shows sample image pairs from our dataset exhibiting different transformation. } 
\end{figure}


\subsection{Description of the PS dataset}{\label{section-PS-qual}}
The PS dataset\footnote{The dataset along with trained models is publicly available at \href{https://github.com/rmitra/PS-Dataset}{https://github.com/rmitra/PS-Dataset}} consists of a total of 30 scenes with 25 scenes for training and 5 scenes for validation. Sample image pairs from the dataset are shown in Fig.~\ref{ps-sample}.  It can be observed from Fig.~\ref{ps-sample} that the diversity of the proposed PS dataset in terms of scene content, illumination, and geometric variations is large.

Each scene in the dataset contains 200 RGB images on an average. The resolution of the images varies from $2000px \times 1350px$ to $800px \times 550px$. The number of patches extracted per scene on an average is 250,000. 
The number of correspondences depend on the threshold imposed on scale and viewpoint variations. 
For the training data used in Sec.~\ref{sec-training}, matching correspondences were obtained by setting scale and viewpoint threshold to $2.5$ and $\left\lbrace 50\degree ,  75\degree \right\rbrace$ respectively. The higher viewpoint threshold is used for scenes which have planar structures.
With these thresholds, on an average, 300,000 matching correspondences per scene are generated. Detailed statistics about each scene is provided in the supplementary material.

\subsection{Creating the dataset}{\label{sec-create-data}}
Structure From Motion (SFM) is adapted to create ground truth pairs of correspondence. To generate the 3D reconstructions, Colmap~\cite{colmap-1,colmap-2} SFM software is used. The SFM process outputs a 3D point cloud with each point having a list of feature points from different images, with which it is triangulated, and predicted intrinsic and extrinsic camera parameters of each image in the scene. Difference of Gaussian (DOG)~\cite{sift} feature points are used in our reconstructions.

Patches are extracted by traversing through the list of feature points associated with each 3D point. An extracted patch is scale and rotation normalized by cropping the patch around the feature point with size $12 \times s$, and then rotating the patch by degree $r$, where $s$ and $r$ are the scale and rotation values of the feature point respectively. The value of $s$ has been limited in the range $\left[1.6, 15\right]$, so that minimum and maximum crop sizes are of $20px$ and $128px$ respectively. The resultant patch is then scaled to $48px \times 48px$. All of the experiments reported in this paper are based on patch size of $32px \times 32px$ which is cropped around the center pixel. This facilitates in avoiding border artifacts when applying data augmentation techniques.

As the PS dataset is constructed from photo collections, there are many instances where a particular scene has images that are captured from almost similar viewpoint and scale. Therefore a sampling technique has been adopted to ensure that the sampled correspondence pairs belonging to a particular 3D point have good coverage of viewpoint and scale. 

\begin{algorithm}
\SetAlgoLined
\KwIn{$p_i$, $\mathbf{p}$, $\mathbf{v}$, $\mathbf{f}$, $\mathbf{d}$}
\KwResult{$m_i$ set of matching correspondences of $p_i$}
 compute matrix A; where A[i][j] contains the angle between $v_i$ and $v_j$\\
 match-found $\leftarrow$ true\\
 \While{match-found $==$ true}{
 	Choose the patch $p_j$ such that,\\
	$j \:=\: \argmax_{k \in \mathbf{p}} \min_{h \in m_i} A[h][k]$ and \\
    $j > i,\: A[i][j] \leq \text{MAX\_V\_TH}$\\ \vspace{2mm}
	
    $MVD_j \:=\: \min_{h \in m_i} A[h][j]$\\  
	$r \:=\: \argmin_{h \in m_i} A[h][j]$\\
    $s_{ij} = \max(f_i / d_i, f_j / d_j) / \min(f_i / d_i, f_j / d_j)$\\
	$s_{rj} = \max(f_r / d_r, f_j / d_j) / \min(f_r / d_r, f_j / d_j)$\\
    \vspace{2mm}
	
    \eIf{$(MVD_j \geq \text{MIN\_V\_TH} \: || \: s_{rj} > 1.5) \: \&\& \: s_{ij} < \text{SC\_TH}$}{
   		$m_i \:=\: m_i \cup p_j$\\
   	}{
   		match-found \:=\: false\\
  	}
}

     \caption{Algorithm to sample matches for patch $p_i$ from $\mathbf{p}$ having suitable scale and viewpoint variation.}
\label{match-pair}
\end{algorithm}

\begin{figure}[h!]
\includegraphics[width=\linewidth]{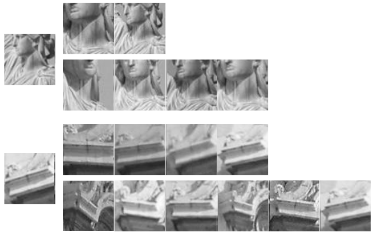}
\caption{\label{view-sample} Examples of sampled patches. The left-most column shows two reference patches. For each reference patch, the matching set in top row and bottom row is generated with MAX\_V\_TH = $40\degree$ and $100\degree$, respectively.}
\end{figure}

\begin{figure*}[t]
\hspace{0.5cm}
\includegraphics[width=7.5cm, height=5cm]{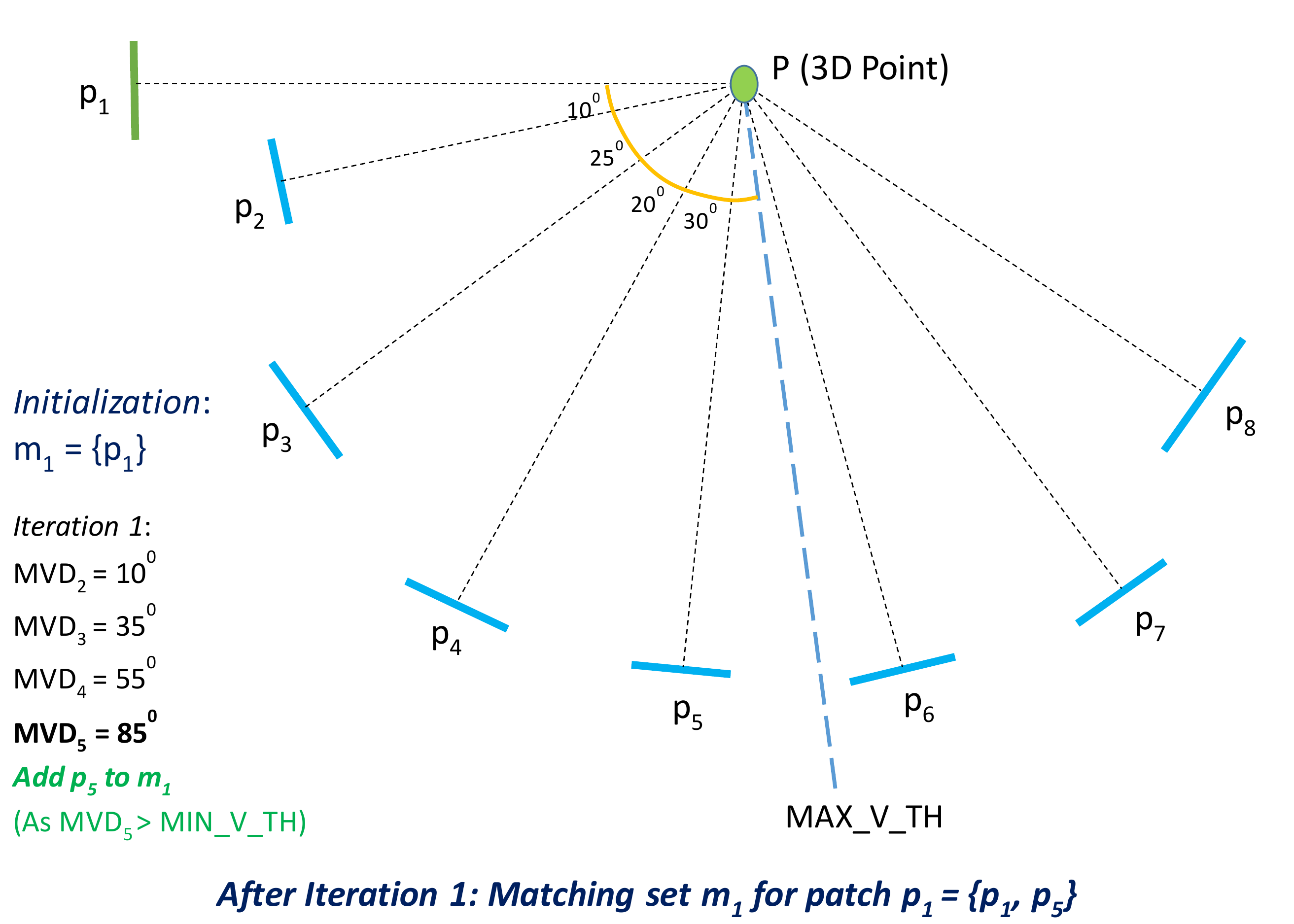}
\hspace{1cm}
\includegraphics[width=7.5cm, height=5cm]{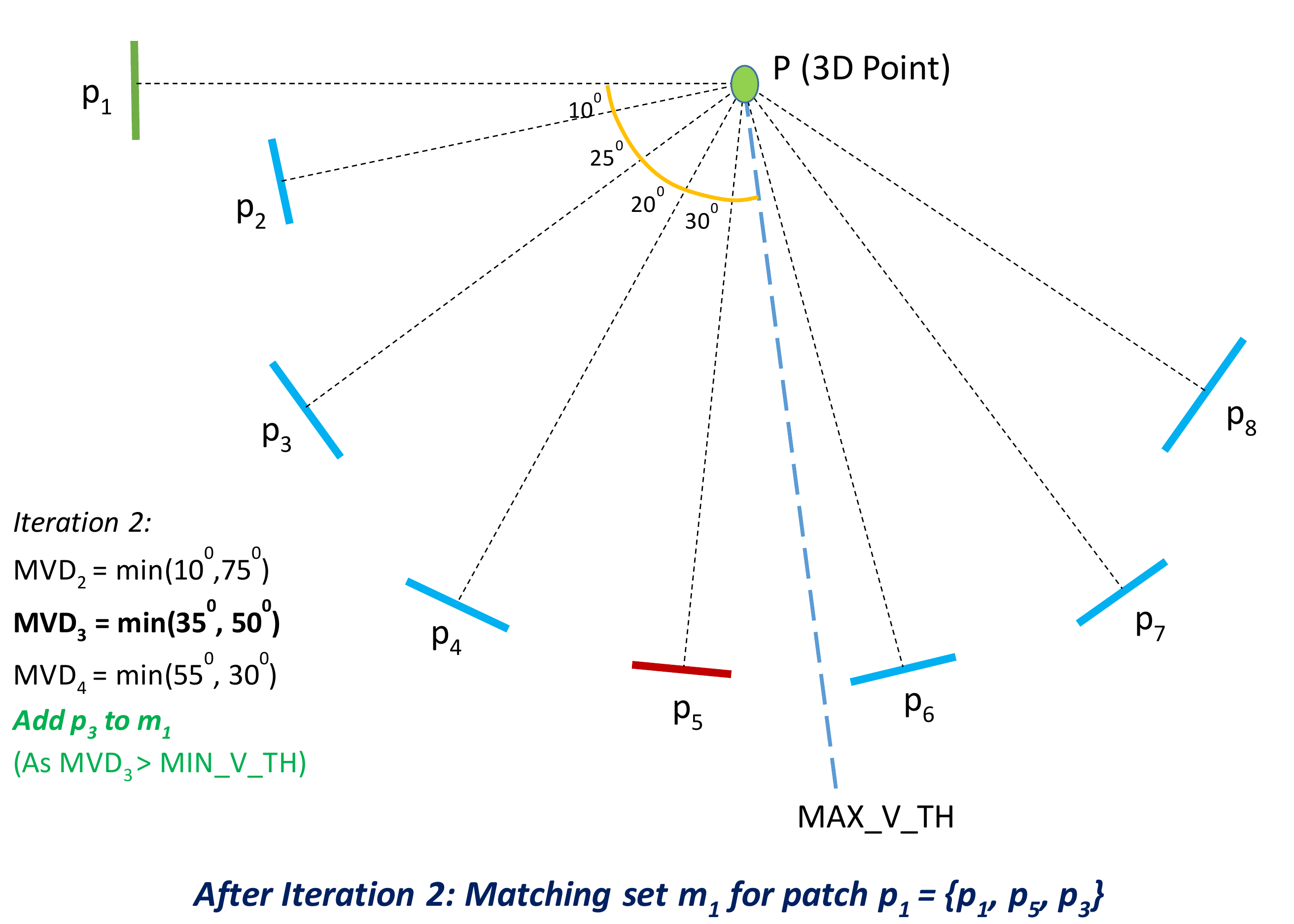} \\

\hspace{0.5cm}
\includegraphics[width=7.5cm, height=5cm]{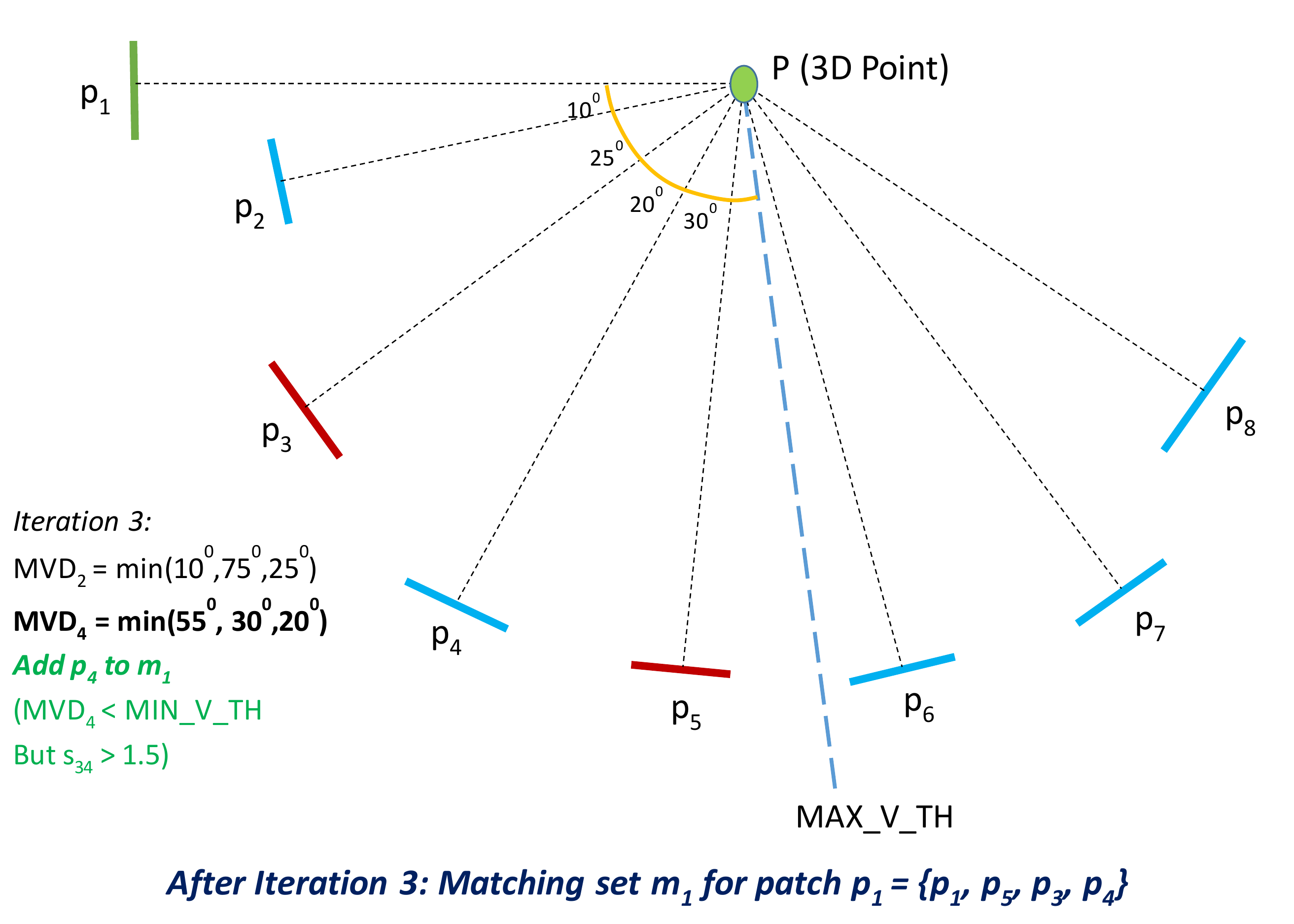}
\hspace{1cm}
\includegraphics[width=7.5cm, height=5cm]{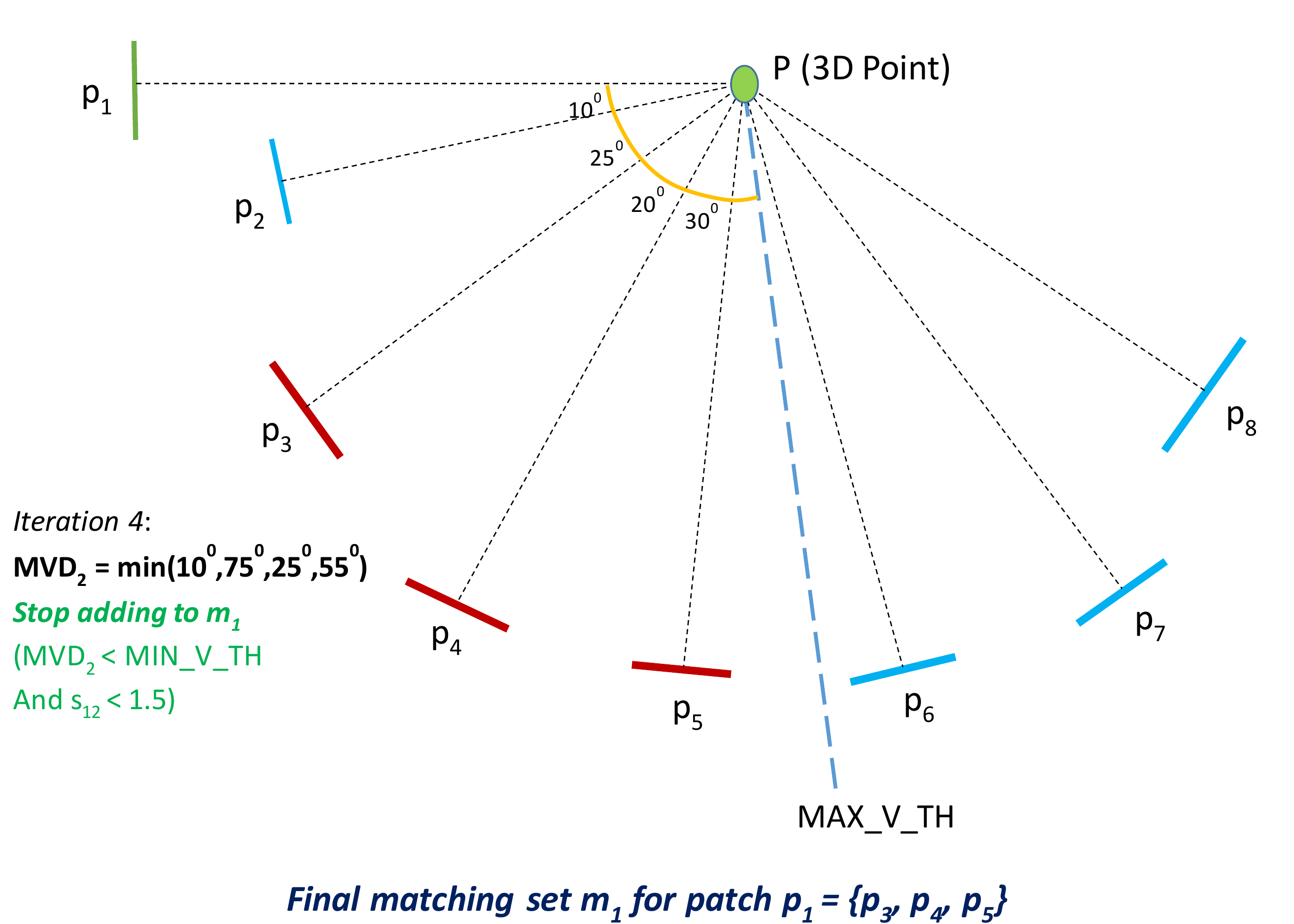}
\caption{An example of sampling technique for identifying matching pairs is shown (figure not to scale). 8 patches $p_i$ of 3D point $\mathbf{P}$ are considered. Here MIN\_V\_TH and MAX\_V\_TH are $25\degree$ and $100\degree$ respectively. Figure shows the iterations for generating the matching set for reference patch $p_1$ (in green). Patches beyond MAX\_V\_TH from $p_1$ are not considered. Each sub-figure shows the patches (apart from $p_1$) in matching set, before start of that iteration, in red. The patch with maximum MVD (in bold) is considered in every iteration. The algorithm stops when no patch is added to the matching set.}
\label{view-sampling-iter}
\end{figure*}

\subsection{Sampling matching correspondences}{\label{sec-sampling}}
Let $\mathbf{P}$ be a 3D point and $\mathbf{p} = \left\lbrace p_1, p_2, \cdots, p_n \right\rbrace$ be the set of patches associated with $\mathbf{P}$. Let $\mathbf{f} = \left\lbrace f_1, f_2, \cdots, f_n \right\rbrace$ be the estimated focal lengths and $\mathbf{v} = \left\lbrace v_1, v_2, \cdots, v_n \right\rbrace$ be viewing directions of cameras of $\mathbf{p}$. Let $\mathbf{c} = \left\lbrace c_1, c_2, \cdots, c_n \right\rbrace$ be the camera centers. We calculate $\mathbf{d} = \left\lbrace d_1, d_2, \cdots, d_n \right\rbrace$ to be the distance of $\mathbf{P}$ from camera centers $\mathbf{c}$ in the direction of $\mathbf{v}$ i.e. $d_i = v_i \cdot (\mathbf{P} - c_i)$. The scale between two patches can be estimated by comparing their $f/d$ ratio. Let SC\_TH, MIN\_V\_TH, MAX\_V\_TH be user defined thresholds for scale, minimum viewpoint difference and maximum viewpoint difference between the pairs. To form matching correspondences with varied viewpoint changes, we initially compute the angle between all possible pairs from $\mathbf{p}$. Next, given a patch $p_i$, its matching set $m_i$ is initialized by $p_i$. Algorithm~\ref{match-pair} has been used to fill the matching set $m_i$. 


The algorithm works in an iterative approach. In each iteration, a patch $p_k$ in $\mathbf{p}-m_i$ and within MAX\_V\_TH from $p_i$, is assigned a minimum viewpoint difference (MVD) value. The value for $p_k$ is computed as follows. The pairwise viewpoint differences (or angles) between $p_k$ and all patches in $m_i$ are computed and the minimum of these differences is assigned as the MVD for $p_k$ in that iteration. This is repeated for all remaining patches in $\mathbf{p}-m_i$ and within MAX\_V\_TH from $p_i$. The patch $p_j$ in $\mathbf{p}-m_i$ having the highest MVD in that iteration is considered. The patch $p_j$ is added to the set $m_i$ if angle between $p_i$ and $p_j$ is more than MIN\_V\_TH or the scale between the two patches differs by at least 1.5. The iterations stop when the algorithm fails to add a patch  to the set $m_i$ in an iteration. The sampling technique avoids adding redundant pairs to $m_i$ which are very similar to already existing pairs. Hence we can obtain the required coverage in viewpoint and scale without creating all possible pairs. Once $m_i$ is computed, patches in the set $m_i \sim \left\lbrace p_i \right\rbrace$ is paired with $p_i$ forming valid matching correspondences. 


An example of sampling method for identifying matching pairs is portrayed in Fig.~\ref{view-sampling-iter} and few examples of matching pairs obtained using the sampling technique is shown in Fig.~\ref{view-sample}.

\section{Experimental Setup}{\label{sec-exp-setup}}
Details of experimental setup used for evaluating various models are discussed in this section. Sec.~\ref{sec-training} gives the detail about procedure followed to train the model using proposed PS dataset. Description about evaluation is given in Sec.~\ref{sec-eval}.
\begin{figure}[h!]
\includegraphics[width=\linewidth]{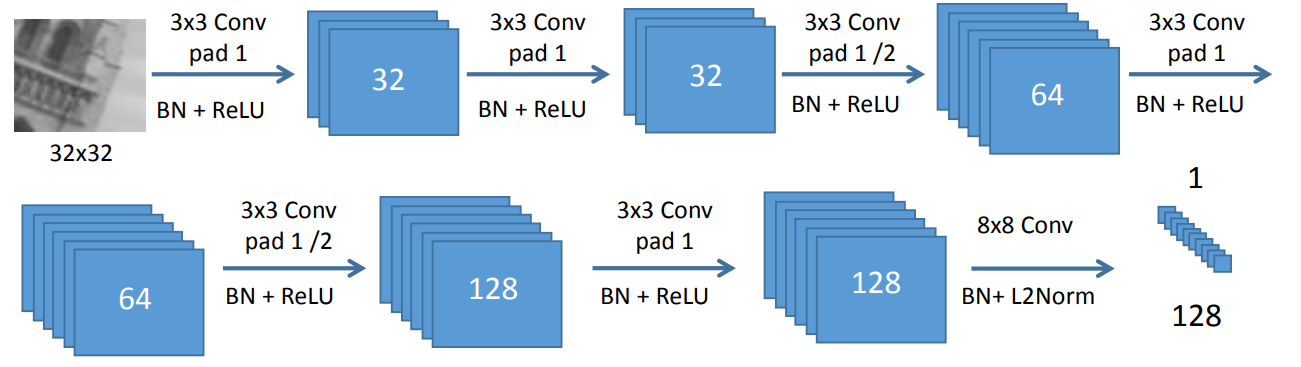}
\caption{The architecture of the network used for training and evaluation. It is the same as the one used by HardNet~\cite{hardnet} without any dropouts. Each convolutional layer is followed by batch normalization and ReLU, except the last one. Similar to HardNet, convolutions with stride 2 are used, instead of pooling in the $2^{nd}$ and $5^{th}$ layer.}
\label{hardnet-arch}
\end{figure}

  
\subsection{Training Procedure}{\label{sec-training}}
For training purpose, the CNN architecture is adapted from \emph{Hardnet}~\cite{hardnet} (also \emph{L2-net}~\cite{l2-net} has similar architecture). Since, the CNN is trained on proposed PS dataset, we call it as HardNet-PS. Schematic diagram of the CNN architecture is shown in Fig.~\ref{hardnet-arch}. It should be noted that the original HardNet and its better variant HardNet+ are trained on MVS dataset~\cite{mvs-new}. 


For comparison with HardNet+, the same loss function as described in~\cite{hardnet} is adapted. In each iteration, $m$ unique 3D points were randomly sampled, where $m$ is the batch size. For a 3D point $\mathbf{P}$ if there are $\mathbf{p} \:=\: \left\lbrace p_1, p_2, \cdots, p_n \right\rbrace$ patches then the hardest from all the $m_i$'s (see Sec.~\ref{sec-create-data}) are chosen based on descriptor distance. Selecting matching pairs from 3D points gives a list of matching pairs $\mathbf{\chi} \:=\: (a_i, b_i); \text{for} \: i \:=\: 1 \cdots m$. 
Next, a pairwise distance matrix $D$ is formed of size $m \times m$, where $D(i, j) \:=\: dist(a_i,\: b_j)$ and function $dist()$ is the L2 distance between the descriptors of $a_i$ and $b_j$. The selection of the nearest non-matching pair $a_{jmin}$ of $a_i$ and $b_{kmimn}$ of $b_k$ are modified as follows:
\begin{eqnarray*}
a_{jmin} \:&=&\: \argmin_{j\: \in valid_{a_i}} D(a_i, b_j) \\
b_{kmin} \:&=&\: \argmin_{k\: \in valid_{b_i}} D(b_i, a_k)
\end{eqnarray*}

where $valid_{a_i}$ contains a set of valid $b_j$'s. Given $a_i$, a patch $b_j$ is valid w.r.t it, when 3D point $\mathbf{P}$ and $\mathbf{Q}$ corresponding to $a_i$ and $b_j$ have at-least one image in common and their projections in that common image differ by $50\%$ of the un-normalized patch size, i.e before scaling to $48$ pixels as done in Sec.~\ref{sec-create-data}. The average loss over the batch is given in Eq.~\ref{hardnet-loss},
\begin{equation}
\begin{split}
	L \:=\: \frac{1}{m} \: \sum_{i\:=\:1,m} \max (0, margin + D(a_i, b_j) \\ - \min ( \: D(a_i, b_{jmin} ), \: D(b_i, a_{kmin} ) )
\end{split}
\label{hardnet-loss}
\end{equation}
To reduce generalization error, augmentation of data is carried out by randomly rotating the patches between $\left[-22.5\degree, \: 22.5\degree \right]$ and scaling within $\left[1.0, \: 1.1 \right]$.  


\subsection{Evaluation procedure}{\label{sec-eval}}
Two evaluation benchmark were used for fair performance comparison, namely, Hpatches for planar objects and Strecha for non-planar objects. The procedure followed to evaluate them are given in Sec.~\ref{sec-hpatch-setup}~and~\ref{sec-strecha-setup} respectively. As in the case with all other descriptors, HardNet-PS is also not trained using any of these two evaluation datasets.

\subsubsection{HPatches Benchmark}{\label{sec-hpatch-setup}}
The \emph{HPatches} benchmark dataset contains image sequences which vary either in viewpoint or in illumination. It has 59 scenes with geometric deformations (viewpoint) and 57 scenes with photometric changes (illumination). Three type of detectors namely DOG, Hessian, and Harris affine are used to extract key points. While extracting key points, additional geometric noise in 3 levels were introduced, namely easy, medium, and hard. Brief overview of the three evaluation procedures or protocols in HPatches are listed below~\cite{hpatches},

\textbf{Patch verification}: Verification is to classify a list of pair of patches as matching or non-matching. Each pair is also assigned a similarity score based on the L2 distance of the descriptors of the two patches. Classification is done on the basis of similarity score. Mean Average Precision (mAP) is calculated based on the list of similarity scores.

\textbf{Image matching}: It is a task of matching key points from reference image to target image. This is done using nearest neighbor on descriptors of the key points. Each predicted match is also associated with a similarity score like patch verification and mAP is calculated over the list of predictions.

\textbf{Patch retrieval}: In this protocol, a patch is queried in a large collection of patches majorly consisting of distractors. A similarity score coherent with the previous evaluations is computed between the query patch and collection of patches. The evaluation is carried out by varying the number of distractors and taking mean.


\subsubsection{Strecha Benchmark}{\label{sec-strecha-setup}}
The \emph{HPatches} benchmark evaluatoin provides a comprehensive evaluation for image sequences related by 2D homography. However, it does not capture image pairs in-the-wild which are non-planar, having self and external occlusions. Hence, the \emph{Herzjesu} and \emph{Fountain} scenes from~\cite{strecha} which have wide-baseline image pairs on non-planar objects has been adapted to evaluate critically. The dataset provides images with projection matrices and a dense point cloud of the scene. The \emph{Herzjesu-P8} scene contains 8 images indexed from 0 to 7 with gradual shift in viewpoint when iterated in order. In other words, the image pair $\{0, 7\}$ has the highest viewpoint difference. Similarly, the \emph{Fountain-P11} scene has a sequence with 11 images.


To ensure high repeatability we assume one of the image in the sequence as the reference image and extract key-points from it and transfer them to the other images. The following steps are used to transfer a point $p_r$ from the reference image to a target image:
\begin{enumerate}
\item Project all 3D points in the reference image.
\item Find the 3D point $P_{nn}$ whose projection onto the reference image is nearest to $p_r$ and within 3 pixels distance.
\item if $P_{nn}$ exists, project it to the other image.
\end{enumerate}
The reference images used in \emph{Fountain-P11} and \emph{Herzjesu-P8} are index ``5'' and index ``4'' respectively.

DOG key-points with 4 octaves and 3 scales per octave were used. The peak and edge threshold are set to $0.02 / 3$ and $10$ respectively. Points with scales larger than 1.6 are retained for stability with at-most 2 orientations per point. \emph{vl\_covdet}~\cite{vl-feat} is used to extract patches from the images with default parameters values. This makes the smallest patch extracted of size $19 \times 19$ which is similar to the support window used by SIFT. In both scenes, we pair all other images with image indexed ``0'' to form the list of image pairs. We divide the image pairs into 3 categories $\left\lbrace \text{narrow},\: \text{wide},\: \text{very-wide} \right\rbrace$ on the basis of viewpoint difference. Range of viewpoint change for ``Narrow'', ``Wide'' and ``Very-Wide'' has been categorized as $\left[ 0\degree\:-\:30\degree \right]$, $\left[ 30\degree\:-\:75\degree \right]$ and $\left[75\degree \:-\:130\degree \right]$ respectively. Table.~\ref{img-pair-list} lists the categorized image pairs of both scenes. Since, \emph{Herzjesu} sequence does not have any image pair differing more than $75\degree$ in viewpoint, the category ``Very-Wide'' is not applicable to it.
\begin{table}[h!]
\centering
\begin{tabular}{||c | P{4em} |P{4em} |P{4em}||}
 \hline
\diagbox[width=8em]{Dataset}{Baseline}
 & Narrow&Wide&Very-Wide\\ [0.2ex] 
 \hline\hline
Fountain-P11  & ``0''-``1'', ``0''-``2'', ``0''-``3'' & ``0''-``4'', ``0''-``5'', ``0''-``6''           & ``0''-``7'', ``0''-``8'', ``0''-``9'' \\
 \hline
Herzjesu-P8   & ``0''-``1'', ``0''-``2'', ``0''-``3'' & ``0''-``4'', ``0''-``5'', ``0''-``6'', ``0''-``7'' & NA\\
 \hline
\end{tabular}
\vspace{2mm}
\caption{\label{img-pair-list} List of image pairs belonging to different baseline categories for the 2 scenes of Strecha dataset. }
\end{table}

Key-point matching is used as metric and followed the same protocol used in HPatches to calculate mAP values. Given a pair of images, we compute the mAP values on 2000 random points visible to both images. 

\section{Results}{\label{sec-results}}

Quantitative comparisons between models trained on MVS dataset and HardNet+ trained on our dataset are described in this section. As described in Sec.~\ref{sec-exp-setup}, Hardnet-PS indicates Hardnet+ trained on proposed PS dataset. Results on Hpatches benchmark evaluation and the Strecha benchmark are discussed in Sec.~\ref{sec-results-hpatches} and Sec.~\ref{sec-results-strecha} respectively.

\subsection{Comparisons on HPatches evaluation benchmark}{\label{sec-results-hpatches}}

Results for matching task are shown in Table~\ref{hpatch-match}. The results are categorized into illumination and viewpoint sequences. As can be observed, in overall score, HardNet-PS outperforms HardNet+ by a margin of 8\%. It is noteworthy that HardNet-PS outperforms all the viewpoint sequences especially on the 'Hard' and 'Tough' sequences by a large margin of 15.5\% and 19.2\%, respectively, over the state-of-the-art.

\begin{table}[h]
\centering
\begin{adjustbox}{max width=\linewidth}
\begin{tabular}{||c|c|c|c|c|c|c|c||}
\hline
\multirow{2}{*}{\diagbox[width=5em]{\small Method}{\small Noise}} & \multicolumn{2}{c}{Easy} & \multicolumn{2}{c}{Hard} & \multicolumn{2}{c}{Tough} & Mean \\
 & Illum & View & Illum & View & Illum & View &  \\
 \hline\hline
SIFT~\cite{sift}       & 43.7 & 49.4 & 18.8  & 21.9 & 08.7 & 09.9  & 25.4\\
\hline
DeepDesc~\cite{deepdesc}   & 42.0 & 46.9 & 23.9  & 27.4 & 13.1 & 14.7 & 28.1 \\
\hline
Tfeat-m~\cite{tfeat}    & 46.7 & 57.6 & 26.3  & 33.7 & 13.9 & 17.4 & 32.7 \\
\hline
Hardnet+~\cite{hardnet}    & {\bf 63.3} & 71.0  & 47.0  & 53.7 & 30.6 & 34.9  & 50.1 \\
\hline
{\bf Hardnet-PS} & 58.6 & {\bf 79.5} & {\bf 49.3} & {\bf 69.2} & {\bf 37.9} & {\bf 54.1} & {\bf 58.3}  \\
\hline
\end{tabular}
\end{adjustbox}

\vspace{1mm}
\caption{Performance comparison for image matching task on HPatches dataset. Illum: illumination sequence. View: viewpoint sequence. Hardnet-PS: Hardnet trained on proposed PS dataset.}
\label{hpatch-match}
\end{table}

The performance comparison on the verification task is shown in Table~\ref{hpatch-ver}. As in the matching task, the sequences can be categorized into same-sequence (intra) and different sequence (inter). Overall, Hardnet-PS is better than Hardnet+ by 4.4\%. The improvement over Hardnet+ increases as the difficulty level of the scenes increase. As it can be seen from Table~\ref{hpatch-ver}, Hardnet-PS performs notably better by nearly 10\% over Hardnet+ on the 'Tough' scenes. 

\begin{table}[h]
\centering
\begin{adjustbox}{max width=\linewidth}
\begin{tabular}{||c|c|c|c|c|c|c|c||}
\hline
\multirow{2}{*}{\diagbox[width=5em]{\small Method}{\small Noise}} & \multicolumn{2}{c}{Easy} & \multicolumn{2}{c}{Hard} & \multicolumn{2}{c}{Tough} & Mean \\
 & Inter & Intra & Inter & Intra & Inter & Intra &  \\
 \hline\hline
SIFT~\cite{sift}        & 86.4   & 79.9  &  67.4   & 59.2   & 52.8   &  44.9   & 65.1 \\
\hline
DeepDesc~\cite{deepdesc}    & 91.3   & 86.2  &  84.0   & 76.2   & 74.2   & 64.8   &     79.5 \\
\hline
Tfeat-m~\cite{tfeat}     & 92.5   & 88.8  &  85.9  &  79.7  &  76.0  &  68.1  & 81.8 \\
\hline
Hardnet+~\cite{hardnet}    &  95.4  &  {\bf 93.0}  &  91.0  &  86.8  &  81.1  &  76.1  & 87.1 \\
\hline
{\bf Hardnet-PS} &  {\bf 95.6}  &  {\bf 93.0}  &  {\bf 94.1}  &  {\bf 90.4}  &  {\bf 90.5}  &  {\bf 85.4}  & {\bf 91.5} \\
\hline
\end{tabular}
\end{adjustbox}

\vspace{1mm}
\caption{Performance comparison for patch verification task on HPatches dataset.}
\label{hpatch-ver}
\end{table}

The results of the retrieval task in the Hpatches evaluation are reported in Table~\ref{hpatch-ret}. The Hardnet-PS outperforms the current state-of-the-art Hardnet+ around 10\% on an average. Again, as in the previous tasks, the margin of improvement for Hardnet-PS is higher for the 'Hard' and 'Tough' scenes by 9.3\% and 16.5\% respectively.

\begin{table}[h!]
\centering
\begin{adjustbox}{max width=\linewidth}
 \begin{tabular}{||c | c |c |c |c||} 
 \hline

\diagbox[width=8em]{\small Method}{\small Noise}
 & Easy&Hard&Tough& Mean\\ [0.2ex] 
 \hline\hline
SIFT~\cite{sift}     	& 64.6 & 37.4 & 22.7 & 41.7 \\
 \hline
 DeepDesc~\cite{deepdesc}   & 67.2 & 52.2 & 37.8 &  52.4\\
 \hline
Tfeat-m~\cite{tfeat}     & 68.4 & 50.8 & 34.7  & 51.3 \\
 \hline
Hardnet+~\cite{hardnet}    & 79.7 & 68.7 & 52.6 & 66.7  \\
 \hline
{\bf Hardnet-PS} & {\bf 82.5} & {\bf 78.0} & {\bf 69.1} & {\bf 76.5} \\
 \hline

\end{tabular}
\end{adjustbox}

\vspace{1mm}
\caption{Performance comparison for patch retrieval task on HPatches dataset.}
\label{hpatch-ret}
\end{table}

\subsection{Comparisons on Strecha Dataset}{\label{sec-results-strecha}}
The mAP values of different models for the matching task on the two datasets of Strecha \etal~\cite{strecha} is shown in Table.~\ref{strecha-fount} and~\ref{strecha-herz}, respectively. Hardnet-PS performs better than the state-of-the-art by nearly 5\% and 3.5\% on the Fountain-P11 and HerzJesu-P8 scenes respectively. The margin of improvement over Hardnet+ is higher in the 'Very-Wide' category for the Fountain-P11 and the 'Wide' category for the HerzJesu-P8 scene.

\begin{table}[h!]
\centering
\begin{tabular}{||c|c|c|P{3em}|c||}
\hline
\diagbox[width=7em]{\small Method}{\small Baseline} & Narrow & Wide & Very-Wide & Mean\\
 \hline\hline
DeepDesc~\cite{deepdesc}    & 76.3  & 40.8 & 9.2  & 40.0\\
\hline
Tfeat-m~\cite{tfeat}     & 86.6  & 62.8 & 21.9  & 57.1\\
\hline
Hardnet+~\cite{hardnet}     & 92.4  & 83.2 & 35.0 & 70.2 \\
\hline
{\bf Hardnet-PS}  & {\bf 92.8}  & {\bf 85.3} & {\bf 47.0} & {\bf 75.0}\\
\hline
\end{tabular}

\vspace{1mm}

\caption{Performance comparison for image matching task on \emph{Fountain-P11} scene}
\label{strecha-fount}
\end{table}

\begin{table}[h!]
\centering
\begin{tabular}{||c|c|c|c|c||}
\hline
\diagbox[width=8em]{\small Method}{\small Baseline} & Narrow & Wide & Mean\\
 \hline\hline
DeepDesc~\cite{deepdesc}    & 64.4 &  13.1 & 35.1\\
\hline
Tfeat-m~\cite{tfeat}     & 76.6 &  27.4 & 48.5\\
\hline
Hardnet+~\cite{hardnet}     & {\bf 85.1}  & 44.5 & 61.9\\
\hline
{\bf Hardnet-PS}  & {\bf 85.1}  & {\bf 50.6} & {\bf 65.4}\\
\hline
\end{tabular}

\vspace{1mm}
\caption{Performance comparison for image matching task on \emph{Herzjesu-P8} scene}
\label{strecha-herz}
\end{table}

\begin{figure}[h!]
\begin{subfigure}[b]{\linewidth}
\centering
\includegraphics[width=\textwidth]{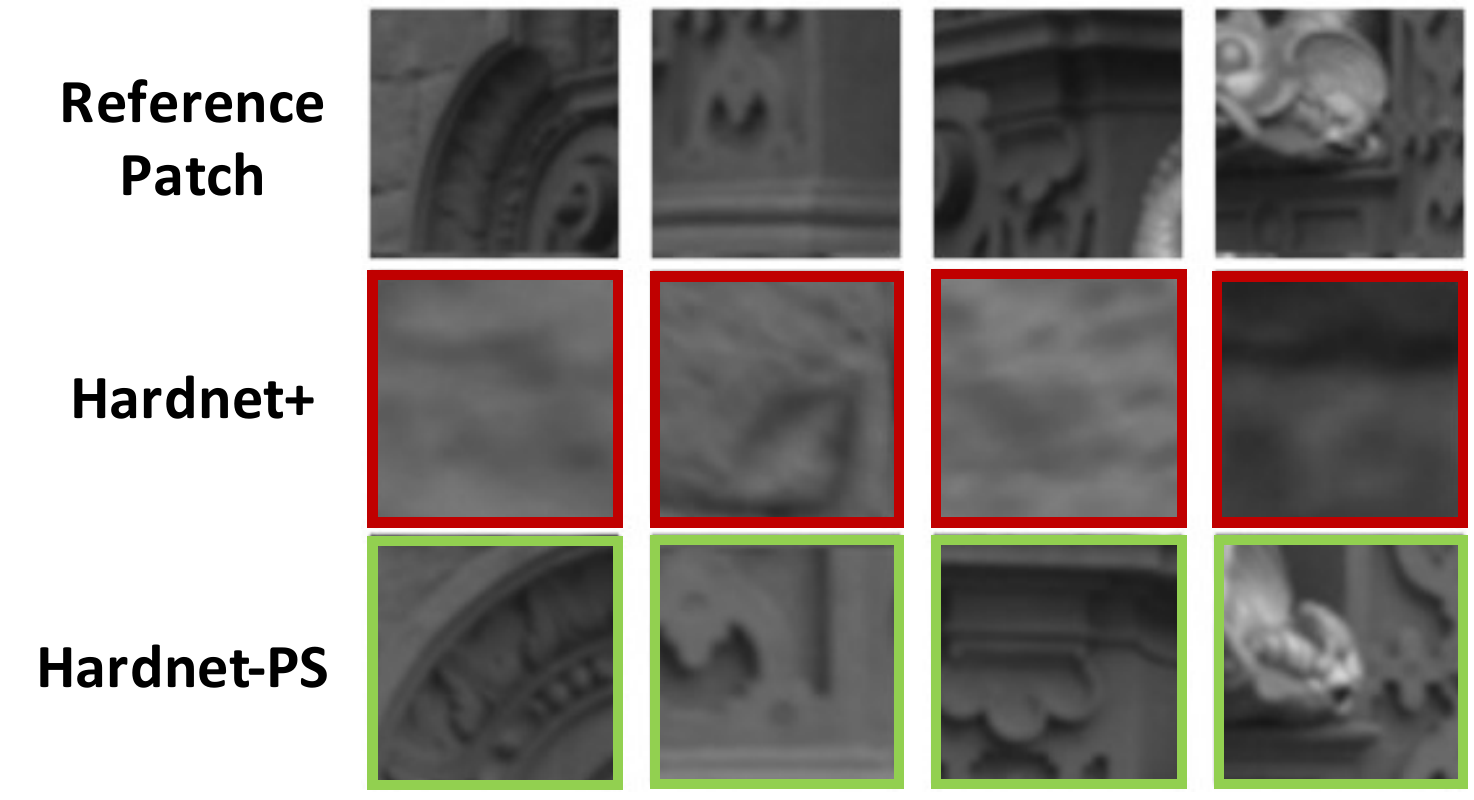}
\caption{Wide Baseline}
\end{subfigure}\\
\vspace{-0.5mm}
\begin{subfigure}[b]{\linewidth}
\centering
\includegraphics[width=\textwidth]{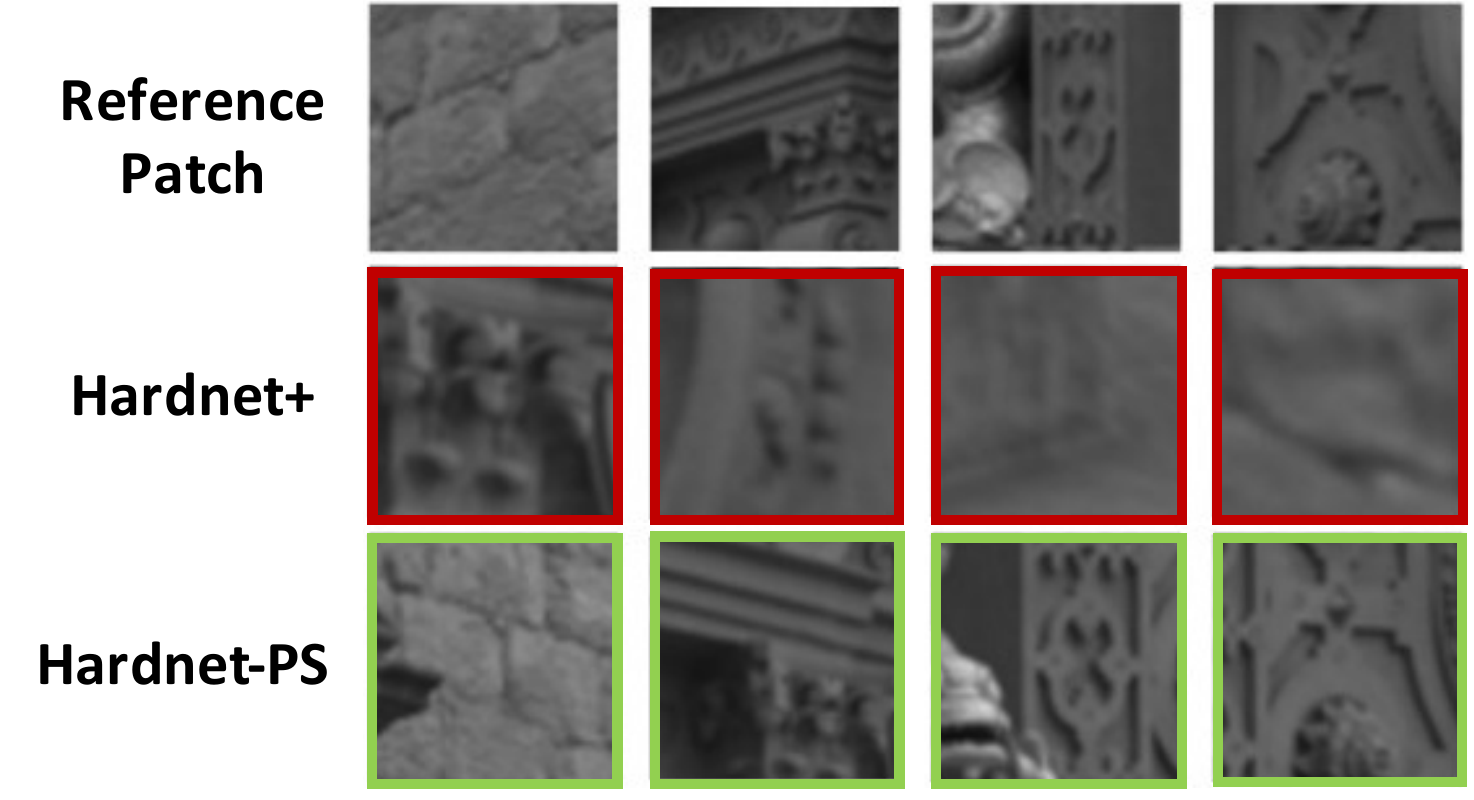}
\caption{{Very Wide Baseline}}
\end{subfigure}
\caption{\label{hardnet-mistake}Examples of incorrect matches made by Hardnet+~\cite{hardnet} while matching ``wide'' and ``very-wide'' image pairs from scene \emph{Fountain-P11}. The top row in both (a) and (b) represent the patches from the source image ``0''. The corresponding predictions are given in the same column. Incorrect and correct predictions are shown in red and green respectively}
\end{figure}

Qualitative comparison for the matching task on the \emph{Fountain-P11} from the Strecha benchmark is shown in Figure~\ref{hardnet-mistake}. It can be seen that for wide baseline and very wide baseline, the matches from the proposed Hardnet-PS model are better than the matches from Hardnet+ model.

The results on the HPatches and the Strecha benchmarks indicate a common pattern. The Hardnet+ and the Hardnet-PS models yield comparably close mAP scores for the 'Easy' scenes (HPatches) and 'Narrow' category (Strecha). But, when the difficulty in the scenes increase ('Hard' and 'Tough' or 'Wide' and 'Very-Wide'), the Hardnet-PS model trained on the PS dataset outperforms the state-of-the-art Hardnet+ model by larger margin.

\section{Conclusion}{\label{section-conclusion}}
In this paper, we have introduced a novel dataset for training CNN based descriptors that overcomes many drawbacks of current datasets such as MVS. It has sufficiently large number of scenes, better coverage of viewpoint, scale, and illumination. We trained the state-of-the-art CNN model available in the literature with the proposed dataset and evaluated on the Hpatches and Strecha benchmark evaluation datasets. On these benchmarks, it has been observed that the model trained with the proposed dataset outperforms the current state-of-the-art significantly, and the margin of improvement is higher for the difficult scenes ('Hard' and 'Tough' in Hpatches and 'Wide' and 'Very-Wide' scenes in Strecha). With these new state-of-the-art results, we conclude that alongside improving the CNN architecture and the training procedure, a good dataset, such as the proposed PS dataset, conforming to the real-world is also necessary to learn high-quality widely-applicable descriptor.

\bibliographystyle{ieeetr}
\bibliography{egbib}

\end{document}